\documentclass[10pt,conference]{IEEEtran} 
\IEEEoverridecommandlockouts
\usepackage[utf8]{inputenc}
\usepackage{algorithmic}
\usepackage{textcomp}
\usepackage{xcolor}
\usepackage{comment}
\usepackage{lipsum}  
\usepackage{hyperref}
    \makeatletter
    \let\NAT@parse\undefined
    \makeatother
\usepackage{graphics} 
\usepackage{epsfig} 
\usepackage{times} 
\usepackage[flushleft]{threeparttable}
\usepackage{cite}
\usepackage{amsmath,amssymb,amsfonts}
\usepackage{url}
\usepackage{dblfloatfix}   
\usepackage{graphicx}
\usepackage{rotating}
\usepackage{multicol}
\usepackage{multirow}
\usepackage[labelformat=simple]{subcaption}
\usepackage{mathtools}
\usepackage[ruled, vlined, linesnumbered]{algorithm2e}
\DeclareCaptionLabelSeparator{periodspace}{.\quad}
\title{\LARGE \bf
Team Mountaineers Space Robotics Challenge Phase-2 \\Qualification Round Preparation Report
}

\author{Team Mountaineers
\thanks{This report prepared by Cagri Kilic, Christopher A. Tatsch, Bernardo Martinez R. Jr, Jared J. Beard, Derek W. Ross under the supervision of Jason N. Gross. Authors are with the Department of Mechanical and Aerospace Engineering, West Virginia University, Morgantown, WV, 3/14/2020}
}

\begin{document}
\maketitle

\begin{abstract}
 Team Mountaineers launched efforts on the NASA Space Robotics Challenge Phase-2 (SRC2). The challenge will be held on the lunar terrain with virtual robotic platforms to establish an in-situ resource utilization process. In this report, we provide an overview of a simulation environment, a virtual mobile robot, and a software architecture that was created by Team Mountaineers in order to prepare for the competition's qualification round before the competition environment was released.  

\end{abstract}

\section{Introduction}
In-situ resource utilization (ISRU) in extraterrestrial soil will allow continuous and affordable human discovery to many deep-space destinations \cite{resourceProspector}. Essential resources on the moon can be used as both vital consumables for humans and building materials of rocket fuel. Moreover, new observations of the moon missions (orbital and surface) have provided evidence of a lunar water system that is more complex and rich than previously believed \cite{colaprete2017resource}, while the proof of lunar volatiles is increasing, the distribution of these resources is not well-known \cite{sanders2012progress}.

NASA is developing an exploration strategy to meet the agency's expanded lunar exploration goals \cite{m2m}. Consistent with this strategy, NASA is planning a series of progressive robotic missions to the lunar surface. To this end, Space Robotics Challenge Phase-2 (SRC-2) was launched by NASA to find solutions to allow a heterogeneous, multi-robot team to autonomously complete tasks envisioned for ISRU. Competitors should develop fully autonomous software to achieve mission tasks in a specified period.

This report provides an overview of Team Mountaineers' virtual planetary simulation used for testing algorithms prior to SRC-2 Qualification Round. Robot Operating System\cite{ROS} (ROS) is the environment in which to test the algorithms with the Gazebo simulator. This made it possible to prototype and test the algorithms quickly in an environment similar to that which is required for the competition. Based on the SRC-2 official rules \cite{SRC2}, the competition will have a Qualification Round and a Competition Round. Both rounds will require fully autonomous operations by virtual robotic systems. The Qualification Round will consist of three tasks;
      \begin{figure}[!htb]
      \vspace{7pt}
      \centering
      \includegraphics[width=\columnwidth]{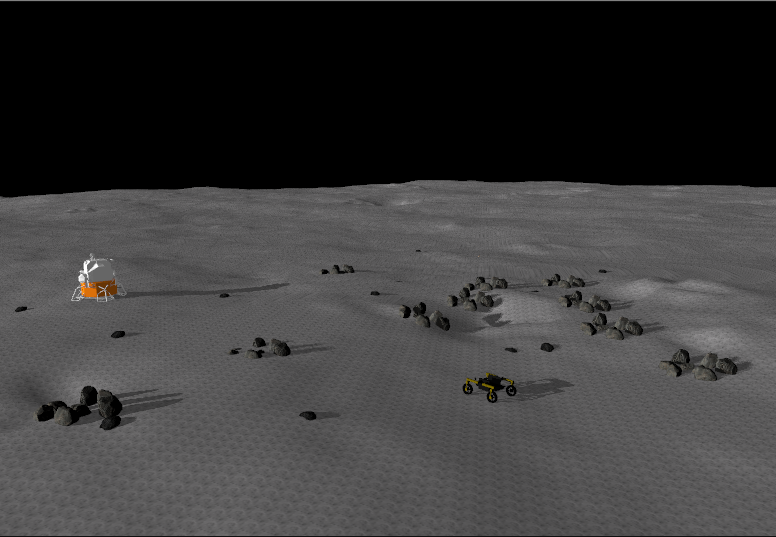}
      \caption{The simulation environment is generated for training purposes using open source Gazebo Lunar terrain. The environment is populated with randomly distributed rocks and our virtual rover (\textit{MoonTamer}) is used in that environment to perform long-term fully-autonomous ISRU mission for the Moon.}
      \label{MoonEnv}
      \end{figure}
\begin{itemize}
    \item Resource Localization: Search the lunar surface with a scout rover for resources and report the location and type.
    \item Resource Collection: Collect a specific amount of resources with a manipulation rover.
    \item Self-Localization: Locate an object, report the location of that object and return to home base. 
\end{itemize}

\section{Simulation}
\subsection{Environment}
The environment is a simulated Lunar surface with the correct gravity value. The environment contains hills, slopes, rocks, and craters. Random distribution is used to populate the rock models in the environment, which can be arranged within a box container with the dimensions given by the user.
\subsubsection{Moon}
In the simulation, the moon environment is adapted from the Open Source Robotics Foundation (OSRF) Apollo 15 Landing Site Gazebo model. The environment uses correct Lunar gravity. The terrain is populated with open source rocks and boulder models\footnote{The rock models can be downloaded from \url{https://3dwarehouse.sketchup.com}} in order to make the environment hard to traverse. The Lunar terrain generation can also be done by using Lunar Reconnaissance Orbiter Camera (LROC) digital terrain model images. The model can be seen in Fig.~\ref{MoonEnv}.


\subsection{Vehicles}
\subsubsection{Base Rover}
\begin{figure}[!h]
      \vspace{7pt}
      \centering
      \includegraphics[width=\columnwidth]{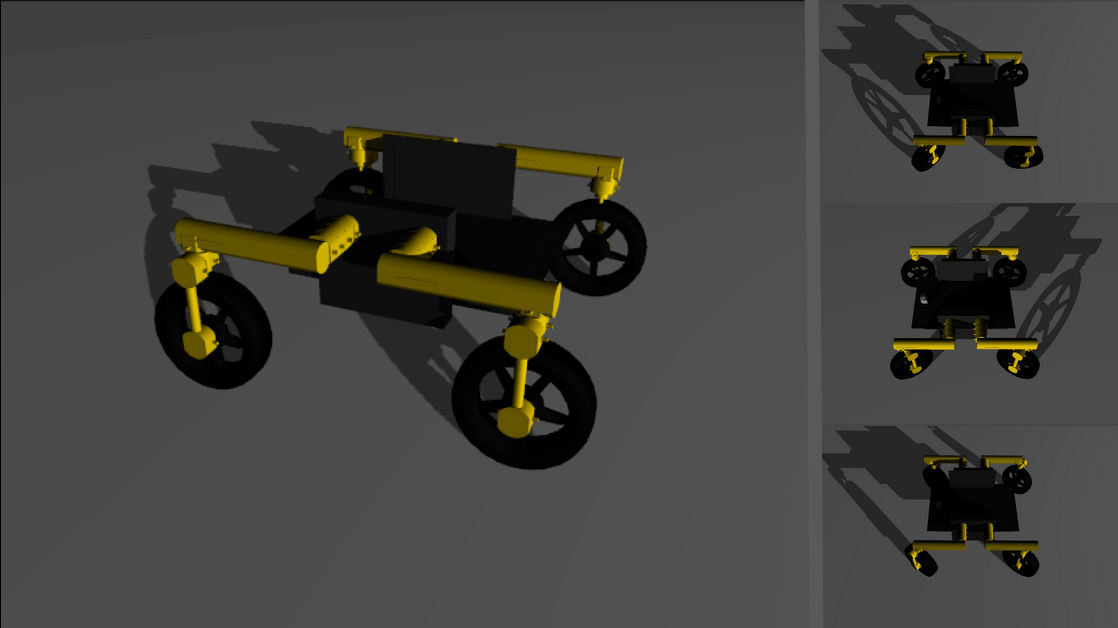}
      \caption{The mobile platform, \textit{MoonTamer}, is generated for testing the algorithms in the simulation environment. MoonTamer is a 4WD rover and able to perform skid-steering, Ackermann steering, omni-directional, and \textit{crab-walk} motion.}
      \label{MoonTamer}
      \end{figure}

Based on the visuals and parameters given from the SRC-2 information packet, Team Mountaineer generated a virtual rover model in Gazebo (see Fig. \ref{MoonTamer}). The rover has four independent wheels that have individual drive motors and are all independently steerable. As such, the rover can be driven in various driving configurations such as Skid-Steering, Ackermann-Steering, and Omni-Directional. 
Wheel velocity can reach 1.5 m/s, and steering can be controlled between -90° and 90° for each wheel. Moreover, the base rover is equipped with several sensors, such as 2D LIDAR, stereo camera, IMU, and wheel encoders.  

\subsubsection{Excavator}

The key to the success of longer space journeys can become independent from Earth supplies; this includes water, propellants, building materials, and many other resources. Therefore, collecting these resources along the way in these long journeys is an essential part of ISRU mission \cite{isru2019}, and it is represented by the Resource Collection task in our context. Collecting materials is a hard task for planetary rovers and includes scooping volatile substances from the terrain. This can be performed with the aid of manipulator arms. There are many examples of manipulator arms being used in space missions; the International Space Station has several manipulators in a few of the recent spacecraft sent to Mars (Mars Polar Lander/Deep Space 2, Mars Exploration Rovers Spirit and Opportunity, Mars Phoenix, Insight Lander, Curiosity Rover, Perseverance Rover) also have them. A beneficial configuration is the one used in The Phoenix Mars Lander Robotic Arm \cite{bonitz2009}. Its robotic arm is a manipulator with four revolute joints that provide motion about shoulder azimuth, shoulder elevation, elbow pivot, and wrist pitch. The end effector has a scooper, a rasp, cameras, and some other sensors that provide information about the extraction of resources.  

\begin{figure}[!htb]
      \vspace{7pt}
      \centering
      \includegraphics[trim={6cm 2cm 3cm 8cm},clip, width=\columnwidth]{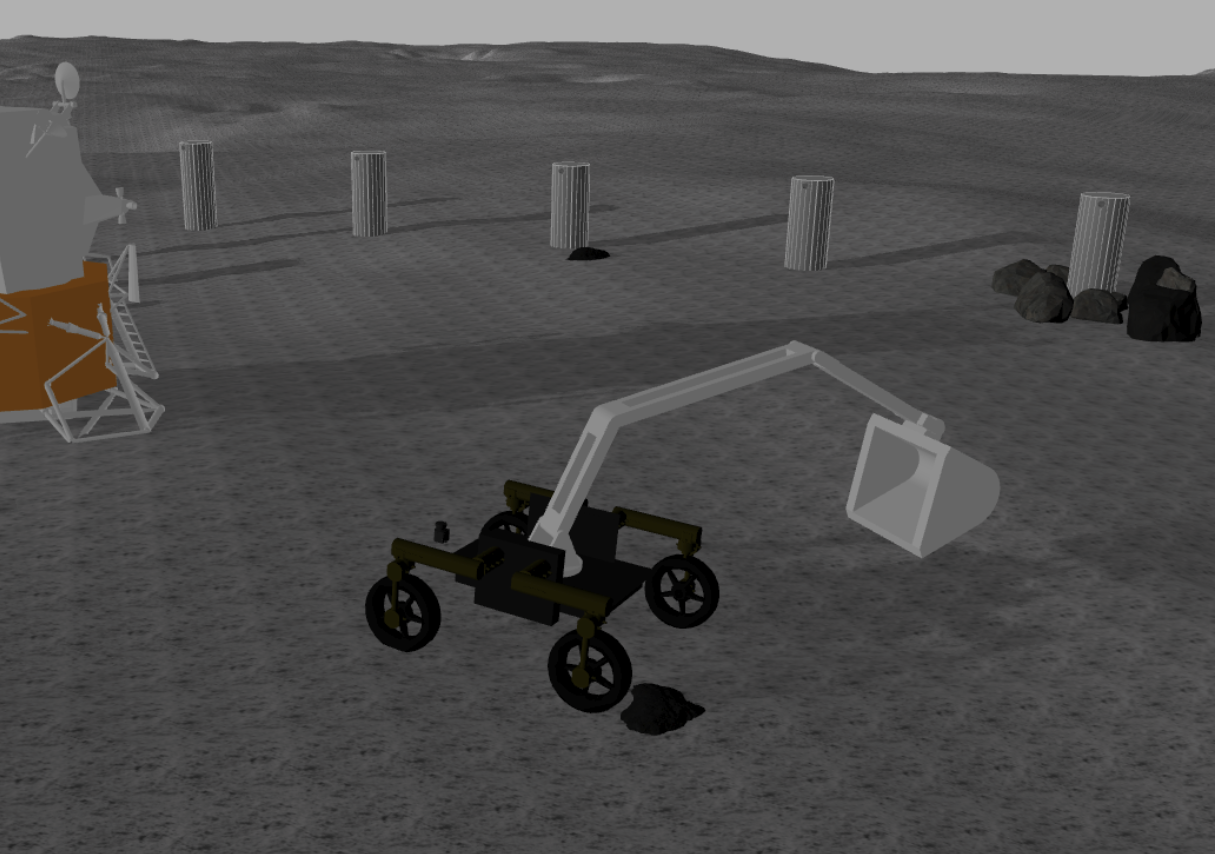}
      \caption{\textit{MoonTamer} Robotic Arm (MTRA): 4R Robot with a Scoop End-Effector (Adaptation of NASA's SRC2 Manipulator).}
      \label{fig:SRC2manipulator}
      \end{figure}

The SRC-2 manipulator arm is a 4R robot (i.e., four revolute joints), very similar to the Phoenix Mars Lander Robotic Arm. The first joint, the shoulder azimuth joint, connects the rover base to the shoulder link, it has no angle limits, and it makes picking up or dropping resources (in this case, the volatile substance) possible from different rover headings. The second joint, the shoulder elevation joint, connects the shoulder with the arm link. This joint has a limited and non-symmetric range of motion. The third joint, the elbow pivot joint, connects the arm with the forearm link. It has a limited and symmetric range of motion. The last joint, the wrist pitch joint, connects the forearm to the bucket link. This joint has a limited range of motion from zero to large angles. It digs and holds volatile at the first portion of the range, and it drops volatile into hauler's bin at the last portion of the angle interval. An adapted version of the robot was designed for our simulator, and it is shown in Fig.~\ref{fig:SRC2manipulator}. 

 \begin{figure*}[b!]
    \centering
    \includegraphics[width=\textwidth,height=0.35\textheight,keepaspectratio]{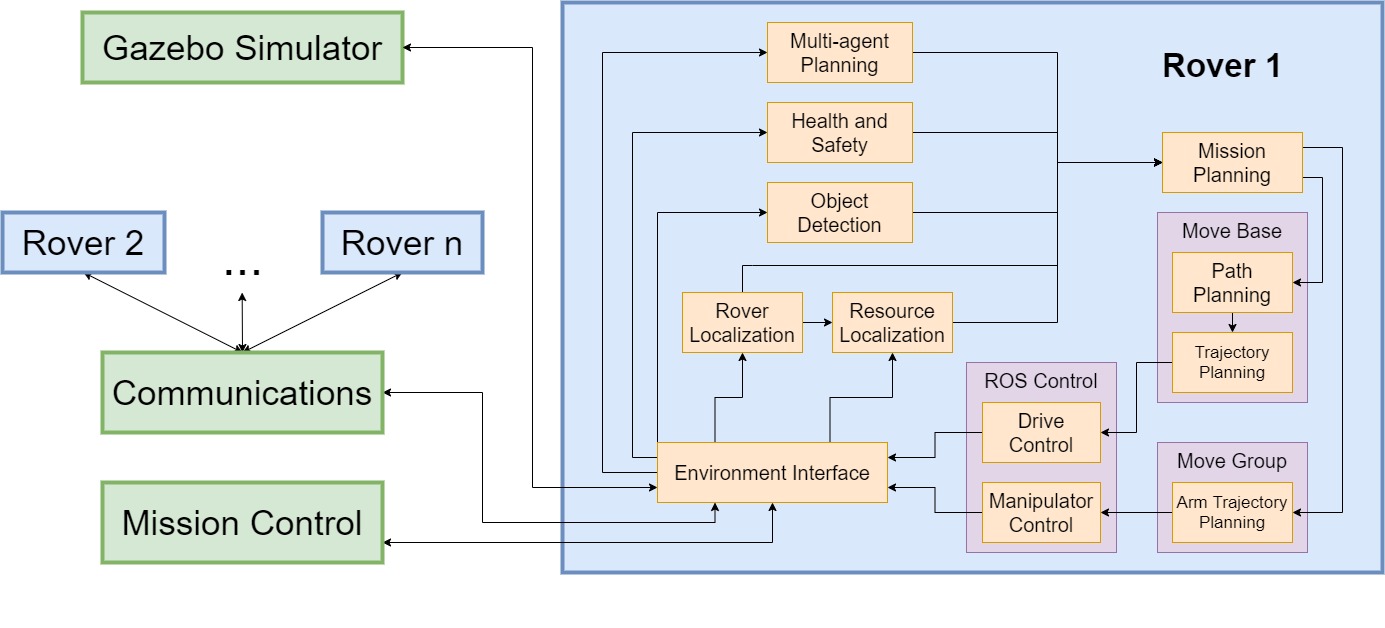}
    \caption{Software architecture diagram}
    \label{architecture_image}
\end{figure*} 

\section{Software Architecture}

\subsection{Overview}

An initial system architecture that we created is detailed in Fig.~\ref{architecture_image} to highlight useful systems and some available packages. An environment interface will bridge the software with its world and the user. This will include behaviors such as communications (whether through ROS or antennae), sensor reading, actuator output, among others. A localization system will ensure the robot understands where it is in the environment. Using the robot's position and ranging information, resources can be located in the environment. The health and status manager will that systems are running as expected and prevent the robot from placing itself in excessive danger. Object detection is charged with identifying salient features such as a Moon lander. Multi-agent processing can provide information with regards to specific objectives and rover allocation. Processing this information through a mission planning node permits the union of potentially disparate objectives and improves decision making. From there, objectives will be translated to lower-level actionable items. This may be facilitated with existing ROS packages such as $move\_base$, $ros\_control$, and $move\_group$ interfaces. 


\subsection{Localization}
Even for a simple task of surface traversing from point A to point B, many decisions need to be made by the autonomous rover to ensure its safe and efficient completion. With the current sensor package, the rover can perform wheel odometry, inertial navigation, and visual odometry (and fusion of these methods). The rover position can be propagated with wheel-odometry when there is no slippage. Stereo vision-based visual odometry can also be used if there are proper lighting and a rich-feature environment. Similarly, the rover hazard detection and avoidance can be performed through processing stereo images and 2D lidar output. It also needs to have precise knowledge of its current and goal positions and maintain a good understanding of what to expect along the way. 
ROS provides several methods to provide localization, which includes types of Kalman filters and particle filters such as ${robot\_localization}$ \cite{MooreStouchKeneralizedEkf2014} which is a collection of state estimation nodes, each of which is an implementation of a nonlinear state estimator for robots moving in 3D space.

\subsection{Mission Planning}
The mission planner is in charge of prioritizing actions and making decisions with respect to how best to meet various objectives (\textit{e.g.}, multi-agent coverage, active perception, digging). Among the most straightforward methods is the implementation of a state machine, which is commonly formulated as a set of conditionals defining the state space. As state machines grow, it can become quite challenging to evaluate the behavior of transitions. To this end, the SMACH state machine package offers a convenient graphical interface to view state transitions \cite{smach}. Furthermore, by breaking each state into its own action server, SMACH allows consideration of each state in a more modular manner. 

\subsection{Path Planning}

Given a goal, path planning is responsible for generating comprehensive paths that the robot is able to follow to reach it. Choosing a path planner depends on several factors, such as robot kinematics and dynamics, type of space (e.g., 3D,2D), sensors available, desired resolution, and update rate. 

For mobile robots, it is common to have multiple levels of planners that connect to each other. 
The $move\_{base}$ \cite{movebase} is a major package from the ROS navigation stack that integrates path planners with sensor data and map information to move a robot to a goal position. 
The core components are $(i)$ $global\_{planner}$ that generates a global 2D path from the start position to goal based on $global\_costmap$, $(ii)$ $local\_{planner}$ that generates a smaller path to a point inside the global path and usually include $local\_costmap$ information and robot dynamics and $(iii)$ $recovery\_behaviors$ that is an emergency action that the robot will take to recover from unexpected a situation. The core package depends on $local\_{costmap}$ that is a 2D map generated directly from sensor data, $global\_{costmap}$, which is a 2D static map that can be imported or generated by robot sensors $odometry$. The  $move\_{base}$ will output $cmd\_vel$ commands to the robot base controller.


Standard global\_planners are  $global\_{planner}$, $carrot\_{planner}$ and  $nav\_{fn}$ which implements A*, Dijkstra and Carrot Planner algorithms to create a global 2D path using the $global\_{costmap}$ data. Writing a custom global planner plugin can be done by writing a new class that adheres to the $nav\_core::BaseGlobalPlanner$, where the $initialization$ and $makePlan$ functions must be re-written with the desired planner. Where the first one initializes the planner name and the costmap2D, and the second takes to start and goal information and generates a PoseStamped vector with the plan.

An interesting example of a custom global planner that adheres to the $move\_base$ framework is the \cite{palmieri2015distance} implementation of rapidly exploring random trees (RRT) and rapidly exploring random trees star (RRT*) where it uses a learning approach to approximate the distance metric for a differential-drive robot configuration.

Standard $local\_planners$ include $DWAPlannerROS$ and $TrajectoryPlanner$ that implement the dynamic window approach (DWA) \cite{fox1997dynamic} and Trajectory Rollout \cite{gerkey2008planning}. DWA is a planner that discretely samples for he robot control space (dx,dy, dtheta) and performs a simulation from the current robot state to predict what happens if different velocities are applied, then each trajectory is scored based on parameters that are defined on the planner and the highest scoring is chosen. Trajectory Rollout is a similar approach but searching for the feasible controls instead of trajectories. Similar to global planners, custom local planners can also be written by adhering to the $base\_local\_planner$ class.

The standard $recovery\_behavior$ is to turn in place. Custom recovery behavior can be implemented by writing a plugin that adheres to the $nav\_core::BaseRecoveryBehavior$.

The standard implementation for all the packages inside the $move\_base$ has many parameters, and tuning them to make it work for a robot can be a challenge, \cite{zheng2017ros} provides a comprehensive guide about most of these parameters and how to tune them.

Move Base Flex \cite{putz2018move} is a similar package that provides more flexibility than the $move\_{base}$ framework. This includes more abstract implementation of Base Local Planner, Base Global Planner, and Recovery Behavior that is not bound to the 2D Costmaps and also more comprehensive results and feedback information during all actions.

To perform well in this challenge requires the ability to find and collect resources in a region as efficiently and accurately as possible. This is closely related to the idea of coverage planning for which a variety of planners have been developed, for which Galceran reviews several methods \cite{galceran2013}. In many cases, the key difficulty decomposing the space into convex polygons or grid cells so a suitable path can be generated. From there, planners will commonly generate a 'lawnmower' pattern within the convex regions. In the case of the grid cell and graph-based approaches, the planners act similar to way-point planners. Coverage planners, do not typically consider the uncertainty involved in search and the lack of an accurate prior map. To this end, Papachristos, $et al.$ performed mapping while mitigating localization errors \cite{papachristos2019}. 

With regard to multi-agent planning, Tully, $et al.$ coordinated motion of the robot's agents in a 'leap-frog' pattern to reduce error growth \cite{tully2010}.


\subsection{Low-level Controllers}

In this simulation environment, there is a critical need for controlling all sorts of robot actuators (e.g., wheels, manipulator joints). A very efficient way of controlling them is using the ROS Control package\cite{ros_control}. This package contains several robot-agnostic controllers, which takes as input the joint state data coming from the simulation environment and the desired state for these joints to produce the desired output that will be fed to the simulator. For this, a generic control loop feedback mechanism, as a PID control, is then used to obtain the output values that will be sent to these actuators. This package has out-of-the-box compatibility with motion planning packages, as MoveIt! \cite{chitta2013}, in case of the manipulation framework, and can be integrated with Gazebo using its plugin for ROS Control \cite{gazebo_control}, using ``transmissions'' and ``hardware interfaces'' in the URDF files. The general purpose controller will accept the position, velocity, or effort control for every single joint in separate control spaces. Nonetheless, the typical use of this package with \textit{MoveIt!} uses a joint trajectory controller as default.

\subsection{Manipulation}

\begin{figure}[!htb]
      \vspace{7pt}
      \centering
      \includegraphics[trim=2cm 0cm 2cm 0cm, width=\columnwidth]{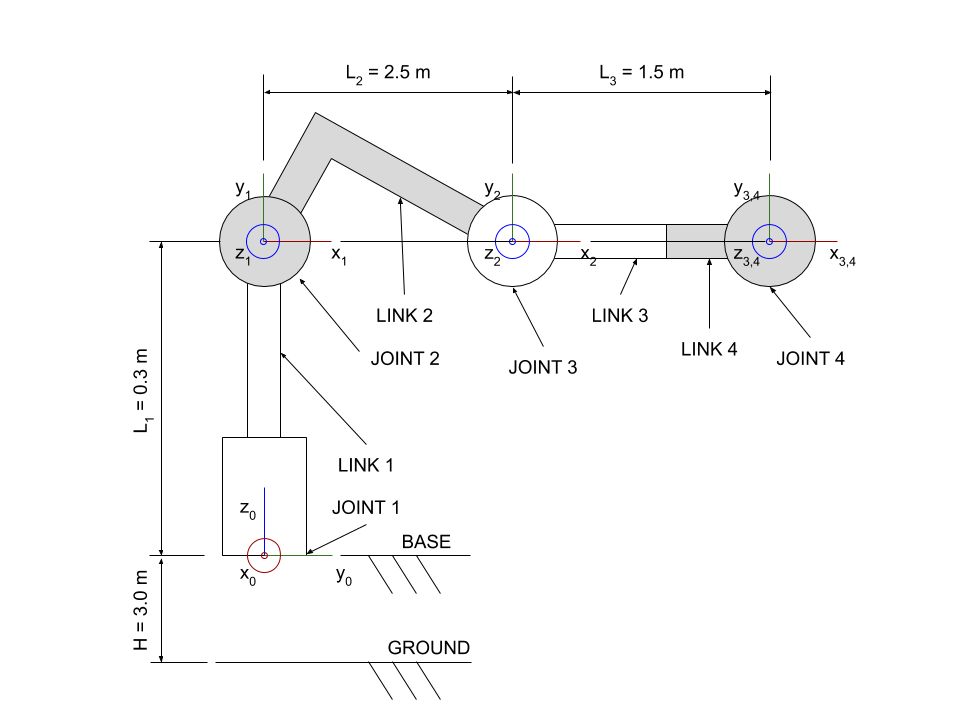}
      \caption{MTRA Main Coordinate Frames.}
      \label{fig:SRC2schematics}
      \end{figure}

The manipulation task of the competition will consist of collecting resources from the terrain. The manipulator has a scoop that will be able to dig volatile substance from below surface level. Manipulator arms can be included in a simulator by adding corresponding links and joints to the rover URDF files. In this case, the manipulator arm in the excavator rover is denoted by the Denavit-Hartenberg Parameters shown in Table \ref{tab:src2mDH}, with corresponding coordinate frames (see Fig.~\ref{fig:SRC2schematics}). Four our simulator, we took the RRBot \cite{gazebo_ros_demos}, three links, two revolute joints arm implemented in Gazebo with extensive documentation, as a starting point. We created an adapted version of the excavator, with dimensions $[l_1, l_2, l_3, l_4] = [0.3, 2.5, 1.5, 0.0]$, generating STL (for collision) and Collada (for visualization) files for each of the links and adding joints and links respecting the DH parameters in the URDF files \cite{gazebo_urdf}. The RRBot only has two joint controllers, so more low-level joint effort controllers were added for the 4R arm, and they were linked to Gazebo. Given the desired paths for the manipulator's arm, it is easy to control the joints using these low-level joint effort controllers by publishing commands in the appropriate topics. 

\begin{table}
\footnotesize
\centering
\begin{threeparttable}
\caption{Denavit-Hartenberg Parameters of 4R Excavator}
\label{tab:src2mDH}
\centering
\begin{tabular}{@{}ccccc@{}}
\hline
Joint i & $a$ [m] & $\alpha$ [rad] & $d$ [m] & $\theta$ [rad] \\
\hline\hline
1 & $0.0$ & $\pi/2$ & $l_1$ & $q_1$  \\
2 & $l_2$ & $0.0$   & $0.0$ & $q_2$      \\
3 & $l_3$ & $0.0$   & $0.0$ & $q_3$      \\
4 & $l_4$ & $0.0$   & $0.0$ & $q_4$  \\
\hline
\end{tabular}
\end{threeparttable}
\end{table} 

 For each of the resources locations, the Mission Planner block (Fig.~\ref{architecture_image}) will provide two important data for the Manipulation Planner block: (1) the resource position with respect to the manipulation base and (2) the hauler position concerning the manipulation base. The trajectory planner will compute a trajectory moving the arm group from its current position to (1), dig the volatile from the terrain, move from (1) to (2), and drop in the hauler. This set of instructions will need to be performed at least twice for each resource location (because the bucket can only carry at most half of the total volatile). This trajectory needs to respect some constraints: maintain the bucket's global angle within a specific range and avoid a collision.

As to our current knowledge, we expect to be able to compute the Forward and Inverse Kinematics of the manipulator offline. Forward Kinematics comes directly from the configuration of the manipulator using the chosen coordinate frames (Fig.~\ref{fig:SRC2schematics}) and its DH parameters. The inverse kinematics is somewhat harder to obtain, but there is enough information to create our solution. For example, \cite{kumar2018} provides a solution for a 3R robot inverse kinematics; this can be used as a first step considering two orthogonal, uncouple planes of motion -shoulder azimuth and shoulder elevation - where latest can be obtained directly from \cite{kumar2018}.
After solving the robot's Forward and Inverse Kinematics, it is possible to obtain its Jacobians and to linearly move the arm in the workspace using by implementing a feedback control in the configuration space. Also, a class of precomputed trajectories respecting can be obtained offline. This guarantees that we will satisfy the constraints (for example, not spilling the volatile after digging), improving robustness while increasing speed. 

Another possibility is to use a ready-to-go ROS package to provide path planning for the manipulator. The MoveIt! package is a compelling and standardized platform for robotic manipulation, which can be easily integrated with the Gazebo simulation environment. Using this package, it is possible to input joint angles and pose goals, obtaining collision aware trajectories. These trajectories are obtained using motion planners libraries, such as the Open Motion Planning Library (OMPL) \cite{sucan2012}. The package plugin ``IKFast'' provides solutions for the robot kinematics, but it also allows changing it for the kinematics given by the user. Finally, the package contains plugins that can check for collisions (if the manipulator and environment meshes, primitive shapes, or Octomaps \cite{octomap} are given) and output trajectories that avoid them. 


\subsection{Computer Vision}

\subsubsection{Visual Odometry}
With each robot having an on-board stereo camera setup, visual odometry (VO) will be an essential capability for rover localization and navigation. Assuming the stereo camera's intrinsic and extrinsic parameters are known, VO can be performed by tracking key points or descriptors between camera frames to estimate the rover's position and orientation relative to its initial position. A basic implementation of VO can be formulated using Random Sample Consensus (RANSAC) for feature matching and built-in OpenCV functions for image processing \cite{opencv_library}. More sophisticated methods will perform Visual-Inertial Odometry (VIO), which takes advantage of the high-rate inertial measurement unit (IMU) in combination with the stereo camera to give a better pose estimate between frames. Fusing the IMU with the camera will also add robustness to localization in dark regions where the camera may not be able to track features effectively. Due to VO requiring moderate-good lighting conditions and an adequate number of features for tracking, the spotlight shall be utilized in these regions if necessary. KIMERA\cite{kimera} and ROVIO\cite{rovio} are two state-of-the-art VIO methods that demonstrate this capability of accurate state estimation in a variety of conditions.

\subsubsection{Object Detection}
One of the challenges of the competition is to detect a small CubeSat and estimate its global position. A potential solution for that is to use neural networks for object detection, localization, and classification. One-stage detectors such as SSD\cite{liu2016ssd} and YOLO\cite{redmon2017yolo9000} and RetinaNet \cite{lin2017focal} allow inference on real-time with high mean Average Precision (mAP) scores. New versions from these network architectures obtain mAP scores of up to 60.6  at 20 FPS on the MS COCO dataset, where the network is trained on over 200 thousand images for 80 object categories. 
These networks can be trained for fewer classes, such as the CubeSat, rocks, base station, other vehicles by performing weight sampling on a previously trained network. Therefore the requirement for a training dataset can be much smaller.

\subsubsection{3D mapping}
The stereo vision sensor is able to provide 3D structural information about the surrounding environment. To generate point cloud data from stereo pair:
\begin{equation}
    u_L = \frac{f(x-d)}{z}, u_R = \frac{f(x+d)}{z}
\end{equation}

\begin{equation} 
(x,y,z) = \left(\frac{(u_L)*z+df}{f}, \frac{(v_L)*z + df}{f}, \frac{2df}{u_R - u_L}\right) 
\end{equation}

where $u_R , v_R$ = camera coordinates in the right camera from robot body frame, $u_L , v_L$ = camera coordinates in left camera from robot body frame, $f$ = focal length of the cameras, $d$ = half the distance between cameras, and $x, y, z$ = coordinate frame between cameras, i.e. camera base frame.
\par
Combining this information with the 2D LIDAR will give more accurate range measurements to obstacles, as shown in \cite{LIDcam}. With the 3D point cloud information, OctoMap \cite{octomap} and RTAB-MAP \cite{rtab} are ROS libraries that can be used to generate an occupancy map and perform Simultaneous Localization and Mapping (SLAM), crucial capabilities for real-time rover path planning.
N 

\section{Conclusions}
\label{conclusions}
In this report, we provide an overview of the preparation efforts of the Team Mountaineers to the SRC-2 Qualification round. A virtual rover, an extraterrestrial environment, and a preliminary software architecture are created and leveraged to test our capabilities. This allowed the team to formulate a list of questions that need to be answered and a testing plan for once the competition simulation is released. 

The reported information and visuals are the outcome of our team's preparation process before the qualification round, and it is not generalizable to other settings or the competition provided software, virtual rovers, and simulation environment.


\addtolength{\textheight}{-2cm}   

\section*{Acknowledgment}

This work was sponsored by the West Virginia University, Benjamin M. Statler College of Engineering and Mineral Resources.

\bibliographystyle{IEEEtran}
\bibliography{references}

\end{document}